\documentclass{article}

\usepackage[preprint, nonatbib]{neurips_2020}

\usepackage[utf8]{inputenc} 
\usepackage[T1]{fontenc}    
\usepackage{hyperref}       
\usepackage{url}            
\usepackage{booktabs}       
\usepackage{amsfonts}       
\usepackage{nicefrac}       
\usepackage{microtype}      
\usepackage{amsmath,xparse}
\usepackage{graphicx}

\usepackage{todonotes}

\title{Explainable 3D Convolutional Neural Networks by Learning Temporal Transformations}

\author{%
  Gabri\"{e}lle ~Ras \textrm{ } Luca Ambrogioni \textrm{ } Pim Haselager \textrm{ } Marcel A.J. van Gerven \textrm{ } Umut G\"{u}\c{c}l\"{u}\\
  Department of Artificial Intelligence\\
  Donders Institute for Brain, Cognition and Behaviour\\
  Radboud University, Nijmegen, the Netherlands \\
  \texttt{$\{$g.ras,l.ambrogioni,w.haselager,m.vangerven,u.guclu$\}$@donders.ru.nl} \\
  }
  
\begin{document}

\maketitle

\begin{abstract}
In this paper we introduce the temporally factorized 3D convolution (3TConv) as an interpretable alternative to the regular 3D convolution (3DConv). In a 3TConv the 3D convolutional filter is obtained by learning a 2D filter and a set of temporal transformation parameters, resulting in a sparse filter where the 2D slices are sequentially dependent on each other in the temporal dimension. We demonstrate that 3TConv learns temporal transformations that afford a direct interpretation. The temporal parameters can be used in combination with various existing 2D visualization methods. We also show that insight about what the model learns can be achieved by analyzing the transformation parameter statistics on a layer and model level. Finally, we implicitly demonstrate that, in popular ConvNets, the 2DConv can be replaced with a 3TConv and that the weights can be transferred to yield pretrained 3TConvs. pretrained 3TConvnets leverage more than a decade of work on traditional 2DConvNets by being able to make use of features that have been proven to deliver excellent results on image classification benchmarks. 
\end{abstract}

\section{Introduction}

Understanding the inner workings of ConvNets is important when they are used to make actionable decisions or when humans have to make actionable decisions based on lower-level decisions from decision support systems. Three-dimensional convolutional networks (3DConvNets)~\cite{baccouche2011sequential} are a natural extension of 2DConvNets and have been investigated for the processing of spatiotemporal data, e.g., action recognition in video~\cite{ji20123d, tran2015learning, varol2017long, carreira2017quo}. While some of these models achieve state-of-the-art results in video action recognition benchmarks, the temporal aspect of their inner workings remains difficult to interpret. Given that a 3DConv filter is simply the 3D extension of the 2DConv filter, one could apply visualization methods that are suitable for 2DConvNets to the individual slices along the temporal axis of a 3DConv filter~\cite{carreira2017quo, anders2019understanding, yang2018visual}. 
This approach is effective to the degree that we can gain insight into what the model learns in terms of spatial features. However, these methods do not provide a meaningful insight into the temporal dynamics that the model takes into consideration. 

Erhan et al.~\cite{erhan2009visualizing} visualized important features in an arbitrary layer of a DNN by optimizing a randomly initialized input such that the activation of the chosen neuron in a layer is maximized. More recent work~\cite{simonyan2013deep, olah2017feature} extends this method, resulting in beautiful visualizations of not only neurons but entire channels, layers, and class representations. One big obstacle to this approach is that the resulting visualizations can be difficult to recognize and are subject to interpretation. Applying these optimization-based methods on 3DConvs will likely exacerbate these problems given that the search space for optimization is at least one order of magnitude larger (depending on the number of frames in the video).

Gradient-based methods reveal which input region causes high activations in specific network components~\cite{simonyan2013deep, montavon2017explaining, zhou2016learning, selvaraju2017grad}. Applying this method to a 3DConv results in a saliency sequence that can be applied to the input video to reveal spatial components that cause a channel to have a high output. Concrete examples can be seen in~\cite{anders2019understanding}, where gradient-based analysis~\cite{simonyan2013deep} and deep Taylor decomposition and layerwise relevance propagation~\cite{montavon2017explaining} are used on a 3DConvNet to analyze model predictions. Yang et al.~\cite{yang2018visual} explain model predictions by adapting CAM~\cite{zhou2016learning} and Grad-CAM~\cite{selvaraju2017grad} for their 3DConvNet. While these methods return an accurate spatial representation of which input components are responsible for high channel output, they are restricted to the spatial domain. 

3TConvNets, in constrast, are 3DConvNets where the convolutional filter is factorized into a 2D spatial filter and corresponding temporal parameters that transform the 2D filter into a 3D filter. The 3TConvNet learns explicit temporal affine transformations whose values can be plotted in an interpretable graph. This alternative approach to learning a 3D filter offers a completely novel way of visualizing and understanding the temporal dynamics learned by a 3DConvNet. In addition, 3TConvs bring the best of both worlds: Because the 3D filter is built from a 2D filter and a set of transformation parameters, many visualization methods can be used as-is or in combination with the transformation parameters. Using 3TConvs gives us a much stronger grasp on the interpretation of spatiotemporal features because the temporal features are directly interpretable regardless of visualization methods used. The goal of this paper is not to improve benchmarks, but to demonstrate that 3TConv can increase human understanding of automated video analyses by providing visual representations of motion features. Thus it contributes to explainable AI and enables a more informed evaluation of the societal implications of AI use cases such as video surveillance.


\section{Related work}
\label{sec:related_work}

Our method bears a resemblance to other types of convolutions that have previously factorized spatiotemporal processing into separate spatial and temporal components and convolutions that incorporate affine transformations. Jaderberg et al.~\cite{jaderberg2015spatial} apply similar affine transformations to features extracted from 2D filters. Our method differs in the sense that it is used to obtain sequentially build a 3D filter, while in~\cite{jaderberg2015spatial} the affine transformations are used to obtain spatially robust feature representations in 2DConvNets.
Group equivariant convolutions~\cite{cohen2016group} make use of symmetry groups containing reflection and rotation to extend translational spatial symmetry. The goal is to learn less redundant convolutional filters in the spatial domain. 
Our method is not the first to apply the concept of separating the spatial and temporal dimensions in 3DConvNets. Tran et al.~\cite{tran2018closer} factorize the individual 3D convolutional filters into separate spatial and temporal components called R(2+1)D blocks. This creates two specific learning phases: a spatial feature learning phase and a temporal feature learning phase guided by one weight per temporal dimension. In our method, we also separate the spatial and temporal components, however, we use four affine parameters instead of one weight. Also, we impose the restriction that the slices in the filter are dependent on one another, thereby avoiding the need for two distinct phases. Instead, spatiotemporal features are learned jointly.


\section{Methods}

\begin{figure}
    \centering
    \includegraphics[width=0.9\linewidth]{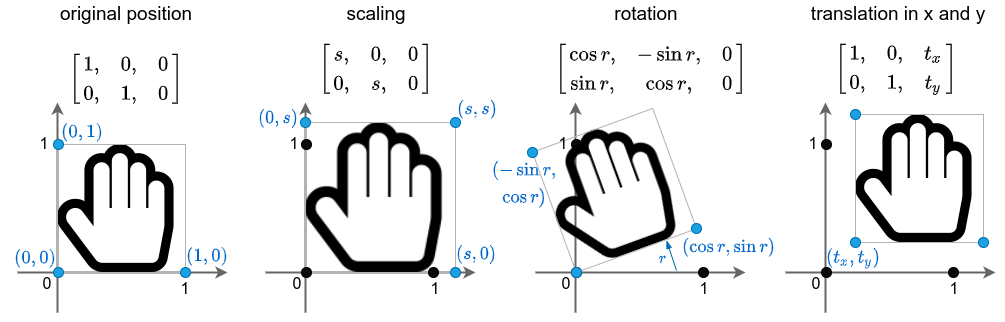}
    \caption{Examples of the affine transformations scale, rotate and translation along the $x$- and $y$-axis, applied on an image of unit height and width. Note that the scaling parameter is equal in both height and width dimensions such that the original aspect ratio is preserved.}
    \label{fig:affine}
\end{figure}

\subsection{Background and notation}
In this section we will introduce some basic notation to describe 3D convolutional layers. We denote the tensor of convolutional filter weights as $\mathcal{K} \in \mathbb{R}^{C \times D \times W \times H}$, where $C$, $D$, $W$ and $H$ refer to the number of channels, the temporal depth, the width and the height of the filter respectively. Slicing $ \mathcal{K}$ along the temporal dimension, we end up with $D$ 2D filters $K \in \mathbb{R}^{W \times H}$. The slice at time lag $t$ is denoted as $K_{t}$. Ignoring the bias term, the forward pass of a 3DConv layer is given by  
\begin{align}
    Y_{a, b, c, s} = \sigma \! \left( \sum_{n,m,k,t} \mathcal{K}_{k, n, m, j} X_{k, a + n, b + m, ,s + t} \right) \,,
\end{align}
where $X$ is the input tensor, $Y$ is the output tensor and $\sigma(x)$ is an activation function. In standard 3DConvNets, each 2D filter $K_j$ is learned independently. This parameterization is arguably unparsimonious as many features of the input stream vary smoothly. In the next section, we will introduce an alternative parameterization in which the $t$-th filter is the result of a differentiable transformation of the $(t - 1)$-th filter.

\subsection{3TConv}
If the input stream has a smooth time dependency, it is reasonable to assume that the filters required to analyze the $t$-th time lag are similar to the filters required to analyze the $t+1$-th lag.\footnote{This is an extension of our earlier preliminary method \cite{anonymous}} We can formalize this insight by parameterizing the $t$-th filter as a smooth transform of its predecessor. I.e.,
\begin{equation}
    K_t = f(K_{t-1}; \Theta_{(t,t+1)})\,,
\end{equation}
where $\Theta_{(t,t+1)}$ is a set of transformation parameters. We denote a filter parameterized in this way as a 3TConv filter.  
Essentially, this is a way to impose a strong sequential relationship between the slices along the temporal dimension of our filter. While regular 3DConvNets learn the entire $\mathcal{K}$ directly, 3TConvNets only learn $K_{1}$ and ${\Theta} \in \mathbb{R}^{(D-1) \times 2 \times 3}$. 
 \subsection{Affine 3TConv}
 One of the main sources of temporal variability in video streams is the optical flow due to the motion of the camera and of the background. These movements induce a constant optical flow that can be modeled as a global affine transformation of the frames. This suggests an affine parameterization of the transformation function in terms of translations, rotations and scaling parameters. Specifically, for every pair of slices $(K_t, K_{t+1})$ we have
\begin{equation}
    \Theta_{(t, t+1)} = 
    \begin{bmatrix} 
        \theta_{11}  & \theta_{12}   & \theta_{13} \\
        \theta_{21}  & \theta_{22}   & \theta_{23}
    \end{bmatrix}  
     = 
    \begin{bmatrix}
        s \cos r & -s \sin r & t_x s \cos r - t_y s \sin r\\
        s \sin r & s \cos r & t_x s \sin r + t_y s \cos r 
    \end{bmatrix}
\end{equation}
Compared to a regular 3D filter which has $W \cdot H \cdot D$ parameters, the 3T filter only has $W \cdot H + 4(D - 1)$ trainable parameters. 

The nonlinear transformation $f(K, \Theta)$ is applied in two stages. First, $\Theta$ is transformed into a sampling grid $\mathbf{G}$ that matches the shape of the input feature map, plus an explicit dimension for each spatial dimension, $\mathbf{G} \in \mathbb{R}^{W \times H \times 2}$.
Here $K_{t}$ is the input feature map and $K_{t+1}$ is the output feature map. 
We should think of this $\Theta \mapsto \mathbf{G}$ transformation as an explicit spatial mapping of $\Theta$ into the input feature space. 
Each coordinate $(x, y)$ from the input space is split in separate $G_x$ with $x \in [1,\ldots, W]$ and $G_y$ with $y \in [1, \ldots, H]$ components, and calculated as
\begin{equation}
    \begin{bmatrix} 
        G_x\\
        G_y
    \end{bmatrix}
=
    \begin{bmatrix} 
        \theta_{11}  & \theta_{12}   & \theta_{13} \\
        \theta_{21}  & \theta_{22}   & \theta_{23}
    \end{bmatrix} 
    \begin{bmatrix} 
        x\\
        y\\
        1
    \end{bmatrix}\,.
\end{equation}
Now that we have sampling grid $\mathbf{G}$ we can obtain a spatially transformed output feature map $K_{t+1}$ from our input feature map $K_{t}$. To interpolate the values of our new temporal filter slice we use bilinear interpolation. For one particular pixel coordinate $(x, y)$ in the output map we compute
\begin{equation}
    K_{t+1, x, y} = \sum_{i=1}^h \sum_{j=1}^w K_{t, i, j} \max(0, 1-|G_x - i|) \max(0, 1-|G_y - j|) \,.
\end{equation}


\subsection{Explaining temporal dynamics with 3TConv}

Temporal parameters ${\Theta}$ containing the scale $s$, rotation $r$ and translation parameters $t_x, t_y$ are obtained from the convolutional layers of a trained 3TConvNet. When a 3TConvNet is first initialized, the parameters are set to the identity mapping: $s=1$, $r=0$, $x=0$ and $y=0$. After training the model, parameter values can be interpreted as follows. The $s$ parameter is the scaling factor relative to 1. The larger $s$, the bigger the resulting transformation. Applied on an image this has the effect of zooming in or out. The $r$ parameter is the rotation measured in degrees where a positive value of $r$ indicates a counter-clockwise rotation. In the resulting plots we multiply $-1 \times r$ such that a positive value indicate a clockwise rotation. The translation parameters in their raw form indicate what percentage of the image has translated. To obtain the amount of translation in pixel units we need to multiply by the width $W$ and height $H$ of the image: $px_x = t_x  W$ and $px_y = t_y H$.


\section{Experiments}
In the following experiments we will demonstrate how 3TConvs can be used for explainability. Modified versions of ResNet18~\cite{he2016deep} and GoogLeNet~\cite{szegedy2015going} are used. In both architectures the 2DConv filters are replaced by 3DConv and 3TConv filters. The networks are trained on the Jester dataset~\cite{materzynska2019jester} to classify 27 different hand-gestures and the UCF101 dataset~\cite{soomro2012ucf101} to classify 101 human activities in various scenarios. Implementation was done in PyTorch~\cite{NEURIPS2019_9015} and for the experiments using transfer learning, model weights are obtained from pretrained GoogLeNet and ResNet18 models from the torchvision model zoo. Details about data pre-processing and model training as well a link to the codebase can be found in the Supplementary Materials.

\subsection{Performance comparison}

Given that the goal of our paper is to demonstrate the explanation capabilities of the 3TConv we only briefly address the performance difference between 3DConvNets and 3TConvNets on the classification of hand- and activity recognition. The performance comparison is shown in Table~\ref{tab:model_results}. On UCF101, pretrained 3TConvNets can outperform pretrained 3DConvNets using 33\% to 53\% fewer parameters. However on the Jester dataset 3DConvNets outperform 3TConvNets. 

\begin{table}[!ht]
\centering
\caption{Classification accuracy and number of model parameters for 3TConv vs. 3DConv.}
\label{tab:model_results}
\begin{tabular}{rcrcr}
\toprule
\multicolumn{1}{c}{} & Jester val acc $\%$ & \# params & UCF101 val acc $\%$ & \# params \\ \midrule
pret. 3T-GoogLeNet & 84.9 & 7419121 & \textbf{63.7} & 7646671\\ 
3D-GoogLeNet & \textbf{89.9} & 14078833 & 58.7 & 14306383 \\ \midrule
pret. 3T-ResNet18 & 74.5 & 11222619 & \textbf{61.1} & 11260581\\
3D-ResNet18 & \textbf{81.6} & 33217755 & 56.1 & 33255717\\
\bottomrule
\end{tabular}
\end{table}


\subsection{Applying 2D visualization methods to 3DConv}
\label{sec:2d_methods_to_3d}

\begin{figure}[!ht]
    \centering
    \includegraphics[width=1\linewidth]{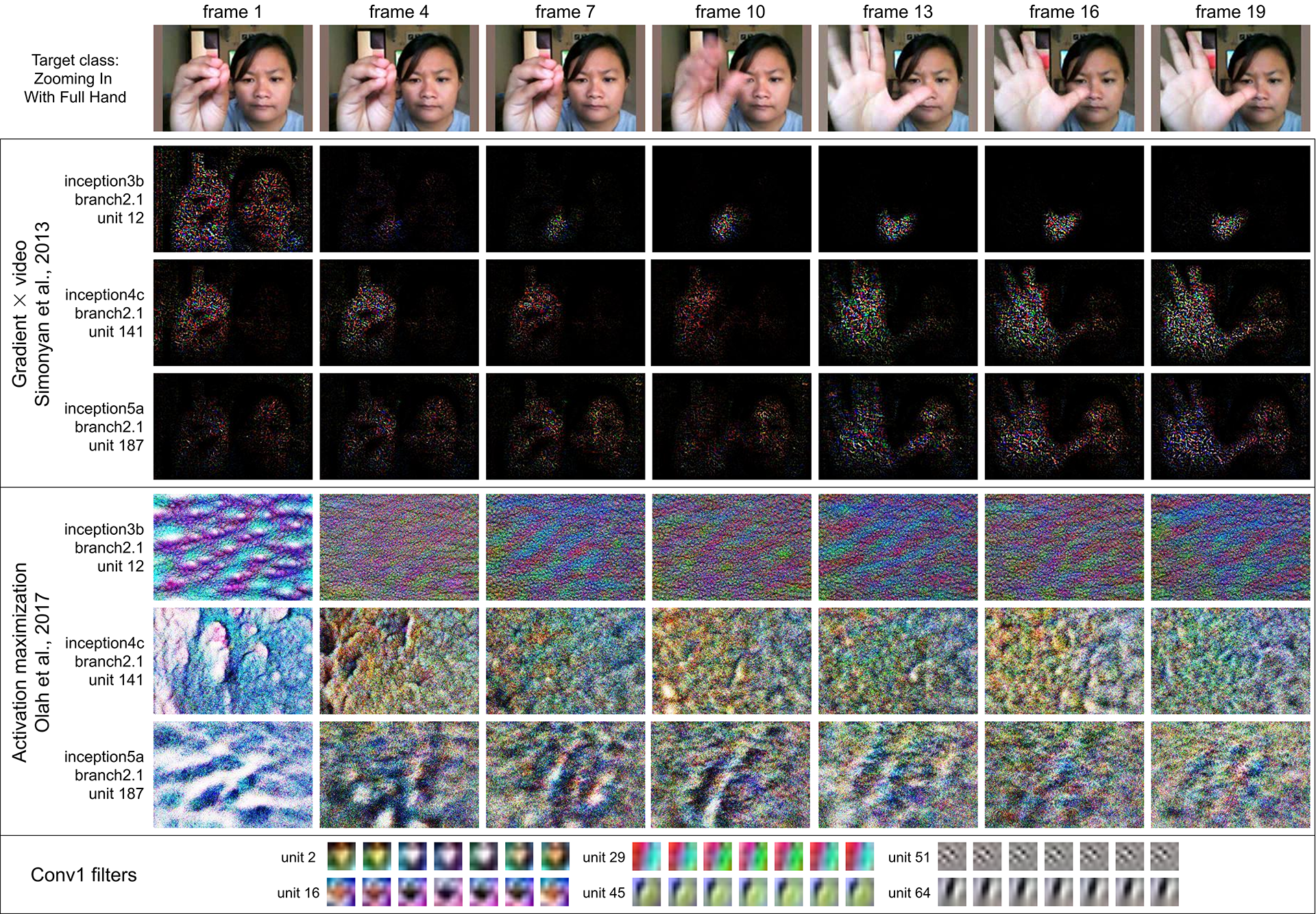}
    \caption{pretrained 3DConv-GoogLeNet trained on the Jester dataset for gesture recognition. Frames from the original video are presented in the top panel. In the subsequent panels, the gradient-based method, activation-maximization and, the visualization of the Conv1 filters are respectively applied to channels of the trained model. These visualizations serve as a baseline.}
    \label{fig:3d_gn_pret_jester}
\end{figure}

Next, we applied the gradient-based method from~\cite{simonyan2013deep} and activation-maximization method from~\cite{olah2017feature} to visualize the Conv1 layer filters directly. The goal is to demonstrate how 2DConv visualization methods translate to the 3DConv-GoogLeNet domain as a baseline for comparing the 3TConv method with.
Figure~\ref{fig:3d_gn_pret_jester} depicts the visualizations. 

The gradient-based method results indicate that the network is picking up spatial components that are both recognizable to humans and that are relevant to classification. That is, areas of the hand are causing specific channels to return high-activation responses. While the method provides an unambiguous indication of the importance of visual components, it does not provide a structure to identify what temporal features are being learned. 

As mentioned in Section~\ref{sec:related_work}, the application of activation-maximization can lead to visual patterns that are difficult to interpret. Figure~\ref{fig:3d_gn_pret_jester} indeed shows that it is difficult to relate the patterns observed in the images to visual aspects of objects in the real world. With no visible continuation between frames, other than the low-frequency patterns themselves, interpretation from a temporal perspective becomes increasingly challenging. The visualizations of the Conv1 filters show that some filters remain unchanged while others exhibit minor color variations. This is likely the result of the small learning rate used to fine-tune the network. Given that the model trains and achieves very reasonable results, we hypothesize that in this model, the spatial features alone are sufficient for the model to make predictions. In any case, no insight can be achieved about the temporal qualities that the model may extract from the data.


\subsection{Using temporal transformations and 2D visualization methods}

\begin{figure}[!ht]
    \centering
    \includegraphics[width=1\linewidth]{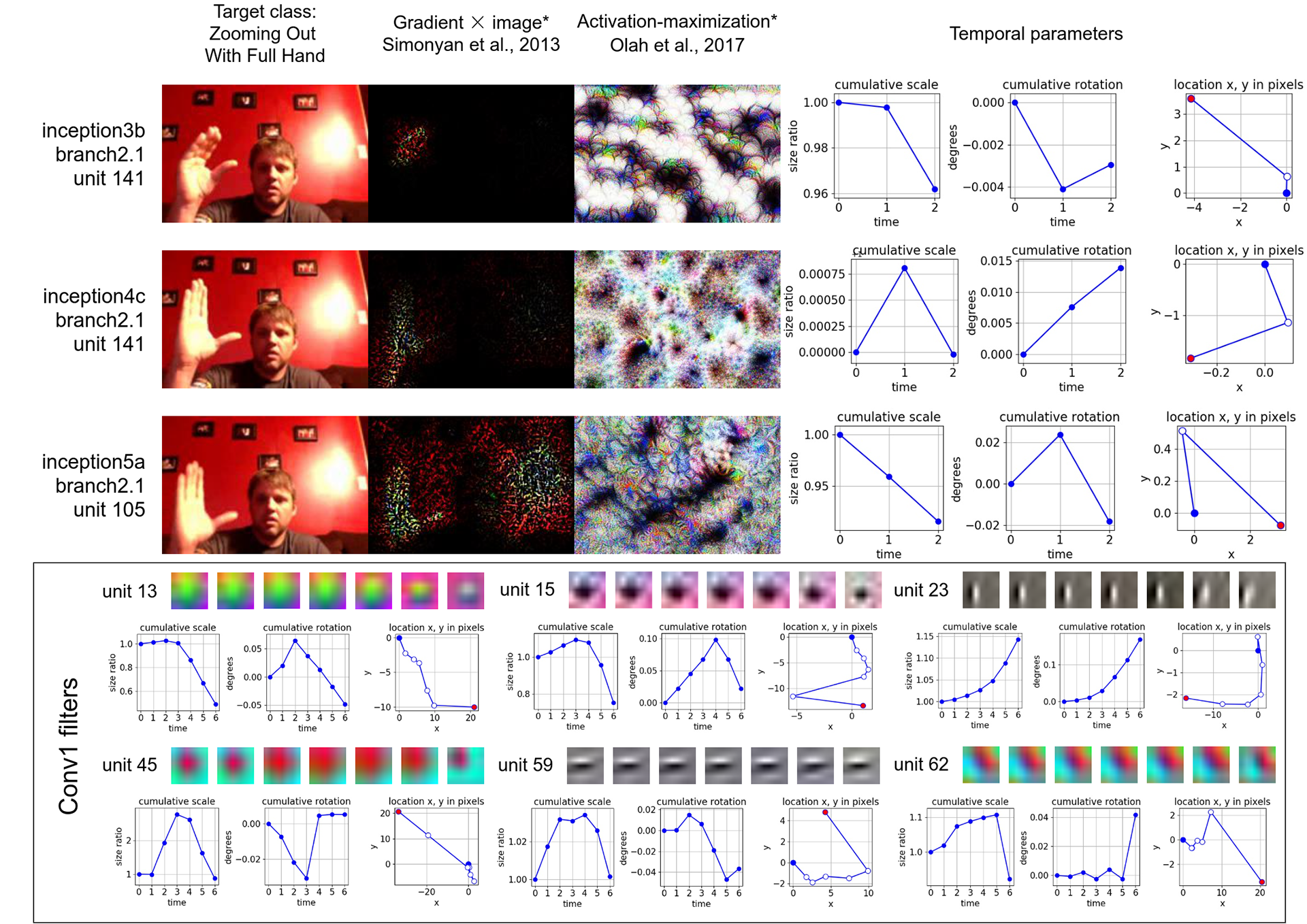}
    \caption{pretrained 3TConv-GoogLeNet trained on the Jester dataset for gesture recognition. In the top panel, from left to right, are displayed the original frame, the result of the gradient-based method, the optimized image for activation-maximization, and the temporal parameter plots of the corresponding channel. In the final column, the location relative to the original position at each timestep is plotted, with the blue marker indicating the starting position and the red marker indicates the ending position. Each row in the top panel represents a unique channel from a specific convolutional layer. In the bottom panel, the visualizations of Conv1 channels are displayed together with their corresponding temporal parameter plots. 
    Modifications were made to the gradient-based method as well as the activation maximization methods in order to produce a single image instead of a series of images.}
    \label{fig:3t_gn_pret_jester}
\end{figure}

To compare the results of 3TConv with 3DConv, the visualization methods in \ref{sec:2d_methods_to_3d} are applied to the pretrained 3TConv-GoogLeNet trained on the Jester dataset. To make use of the transformation parameters, both gradient-based and activation-maximization were adapted to only produce an image and not an entire video.\footnote{Note that this strategy can also be applied to 3DConvNets, however, given that 3DConvs do not have explicit temporal parameters, no insight into the temporal mechanics can be derived.} In the gradient-based method, only the frame causing the highest activation is shown. In the activation-maximization method, a single frame is optimized and stacked on top of each other to create a video input. Results are shown in Figure~\ref{fig:3t_gn_pret_jester}.

Similar to the results for the 3DConv-GoogLeNet, the gradient results for 3TConv-GoogLeNet show that recognizable components in the input are responsible for the high activations in the channels. The visualizations for activation-maximization remain uninterpretable, however, we do notice that the images exhibit more structure compared to the ones in Figure \ref{fig:3d_gn_pret_jester}. This is likely the result of optimizing for a single image rather than optimizing a video; optimizing for a single image is an easier task because an image contains fewer free parameters compared to a video. For the indicated channels in the Conv1 visualizations, we can see transitions that correspond with the temporal parameters. For the channels in each example, we can plot the temporal parameters directly and gain insight into what that particular channel has learned. For example, for inception3b we can directly read from the graphs that channel 141 learned the motion of zooming out while moving up-left and rotating counterclockwise. This provides us a novel way to explain model behavior from the temporal perspective independent from the visualization methods used.


\subsection{Analysis of temporal parameters across models and datasets}

\begin{figure}[!ht]
    \centering
   \includegraphics[width=1\linewidth]{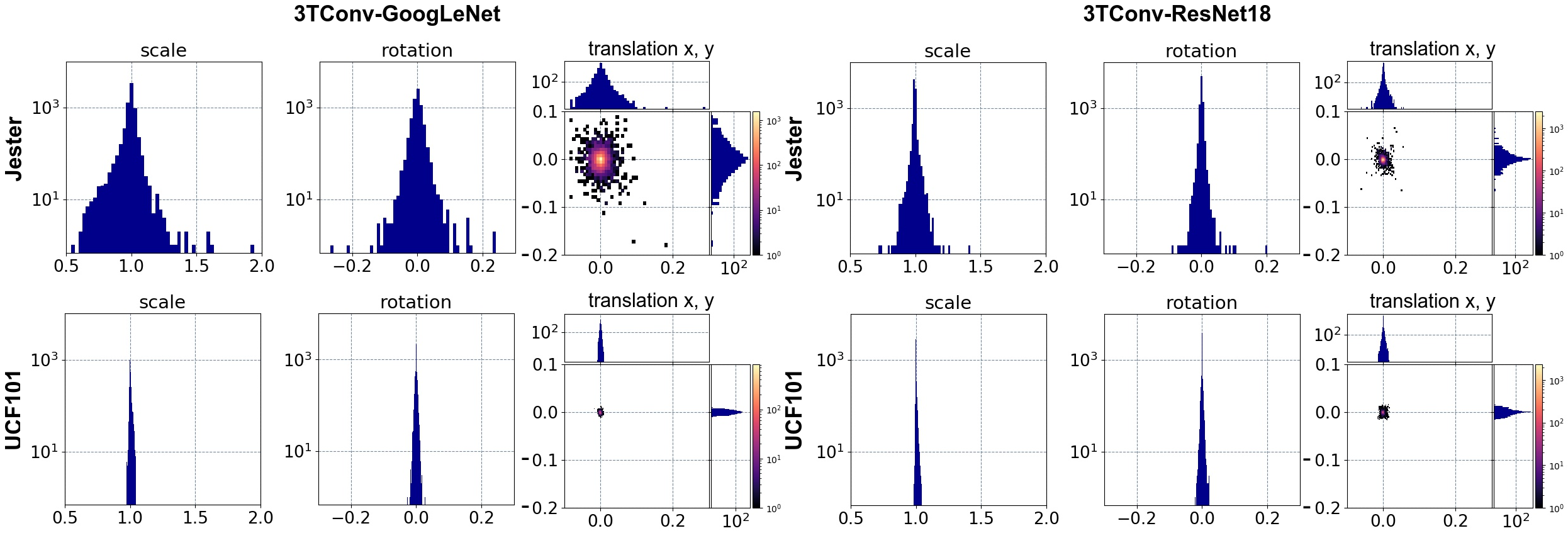}
    \caption{Distributions of the temporal parameters of pretrained 3TConv versions of GoogLeNet and ResNet18. Each network has been trained on the Jester dataset for gesture recognition and on the UCF101 dataset for action recognition. Each separate panel represents the distributions of all the learned temporal parameters across a single model after training on the indicated dataset. In each separate panel, three distributions are displayed: (i) the distribution for the scale parameter, (ii) the distribution for the rotation parameters, and (iii) the joint distribution for the translation parameters. Note that the axis scales are equal for each panel so that a direct comparison can be made, between models and well as between datasets.}
    \label{fig:pret_model_distribution}
\end{figure}

In the previous section, the temporal parameters were analyzed on a per-channel level. The temporal parameters can also be used to understand the global decision-making process of the model in interpretable and quantifiable model statistics. In Figure~\ref{fig:pret_model_distribution}, the distributions of the learned temporal parameters of 3TConv-ResNet18 and 3TConv-GoogLeNet trained on the Jester and UCF101 datasets are visualized.

Figure \ref{fig:pret_model_distribution} is summarized in Table \ref{tab:model_statistics}, where the relative values of each temporal parameter can be captured with by its mean value $\mu$ and standard deviation $\sigma$. These summary statistics allow us to directly compare different models trained on different datasets. We immediately attain a high-level view of what each network considers important for classification for the specific datasets.

\begin{table}[!ht]
\centering
\caption{Means and standard deviations of estimated parameters for pretrained models ($\times 10^{-3}$).} 
\label{tab:model_statistics}
\begin{tabular}{@{}ccccc@{}}
\toprule
\multicolumn{1}{l}{} & \multicolumn{2}{r}{pret. 3TConv-GoogLeNet} & \multicolumn{2}{r}{pret. 3TConv-ResNet18} \\ \midrule
\multicolumn{1}{l}{} & \textbf{Jester} & \textbf{UCF101} & \textbf{Jester} & \textbf{UCF101} \\ 
\multicolumn{1}{l}{parameter} & $\mu$ / $\sigma$ & $\mu$ / $\sigma$ & $\mu$ / $\sigma$ & $\mu$ / $\sigma$ \\ \midrule
$s$ & 28.0 / 46.14 & 6.24 / 7.55 & 8.46 / 16.1 & 3.57 / 5.3 \\
$r$ & 11.11 / 13.77 & 2.38 / 2.67 & 3.26 / 5.44 & 1.65 / 2.55 \\
$p_x$ & 7.58 / 10.27 & 1.47 / 1.48 & 1.87 / 3.16 & 1.49 / 2.07 \\
$p_y$ & 8.14 / 10.75 & 1.71 / 1.65 & 1.99 / 3.5 & 1.51 / 2.21 \\ \bottomrule
\end{tabular}
\end{table}

Results show that models trained on the Jester dataset develop temporal parameters that vary more in value and range compared to the parameters of the UCF101 dataset.

\subsection{Results for tabula rasa models}


Finally, to quantify the benefits of using pretraining, we compare the results of the previous section with those of tabula rasa models that were trained from scratch.

\begin{table}[!ht]
\centering
\caption{Comparison of classification accuracy for pretrained models vs. tabula rasa models that were trained from scratch.}
\label{tab:pret_vs_scratch}
\begin{tabular}{lcc}
\toprule
\multicolumn{1}{c}{} & Jester val acc $\%$ & UCF101 val acc $\%$ \\ \midrule
pret. 3T-GoogLeNet & \textbf{84.9} & \textbf{63.7} \\
scratch 3T-GoogLeNet & 74.1 & 33.4 \\ \midrule
pret. 3T-ResNet18 & \textbf{74.5} & \textbf{61.1} \\
scratch 3T-ResNet18 & 44.4 & 31.0 \\
\bottomrule
\end{tabular}
\end{table}

 Table \ref{tab:pret_vs_scratch} compares the classification accuracies for 3TConvnets that were either pretrained or trained from scratch. Pretrained models massively outperform tabula rasa models, showing that transfer learning has a significant impact on 3TConv performance.


\begin{figure}[!ht]
    \centering
    \includegraphics[width=1\linewidth]{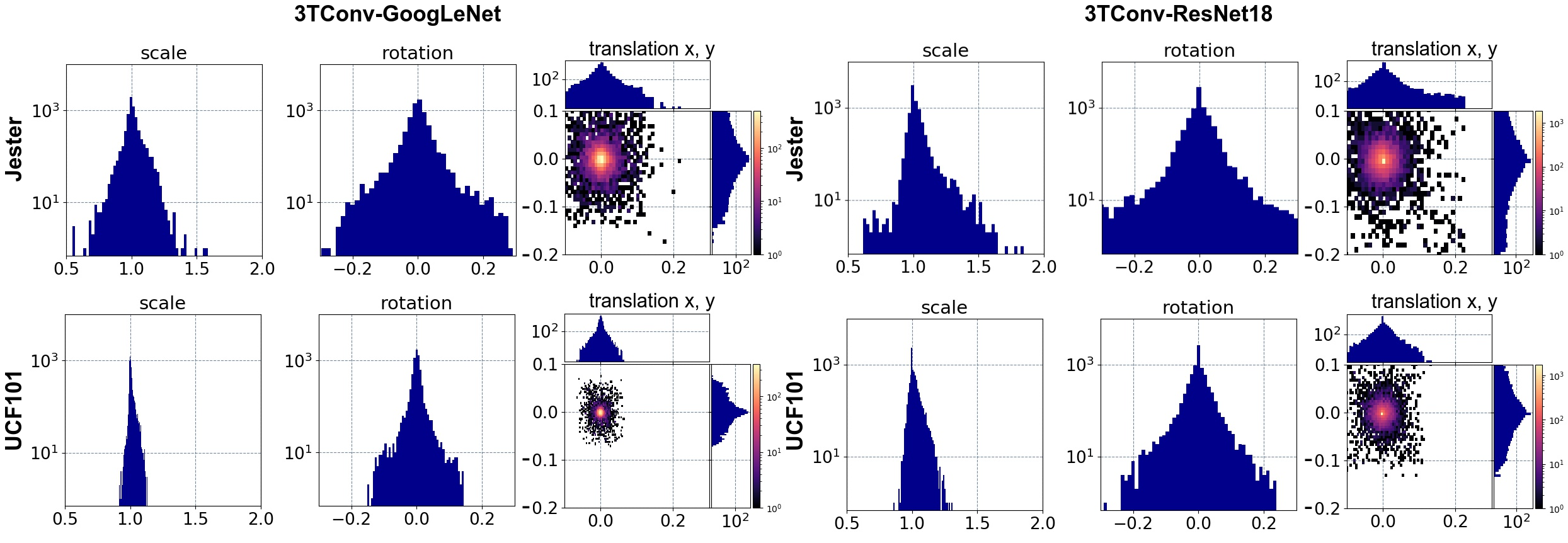}
    \caption{Distributions of the temporal parameters for the 3TConv versions of GoogLeNet and ResNet18 that were trained from scratch. Each network has been trained on the Jester dataset for gesture recognition and on the UCF101 dataset for action recognition. Each separate panel represents the distributions of all the learned temporal parameters across a single model after training on the indicated dataset. In each separate panel, three distributions are displayed: (i) the distribution for the scale parameter, (ii) the distribution for the rotation parameters, and (iii) the joint distribution for the translation parameters. Note that the axis scales are equal for each panel so that a direct comparison can be made, between models and well as between datasets.
    }
    \label{fig:scratch_model_distribution}
\end{figure}

\begin{table}[!ht]
\centering
\caption{Means and standard deviations of estimated parameters for tabula rasa models ($\times 10^{-3}$).} 
\label{tab:model_statistics2}
\begin{tabular}{@{}ccccc@{}}
\toprule
\multicolumn{1}{l}{} & \multicolumn{2}{r}{3TConv-GoogLeNet} & \multicolumn{2}{r}{3TConv-ResNet18} \\ \midrule
\multicolumn{1}{l}{} & \textbf{Jester} & \textbf{UCF101} & \textbf{Jester} & \textbf{UCF101} \\ 
\multicolumn{1}{l}{parameter} & $\mu$ / $\sigma$ & $\mu$ / $\sigma$ & $\mu$ / $\sigma$ & $\mu$ / $\sigma$ \\ \midrule
$s$	&	40.14 / 67.09	&	26.86 / 34.69	&	42.37 / 51.92	&	14.85 / 19.47	\\
$r$	&	30.12 / 47.2	&	23.16 / 32.11	&	29.31 / 38.26	&	13.12 / 19.32	\\
$p_x$	&	19.88 / 32.05	&	13.72 / 17.86	&	18.21 / 22.85	&	7.06 / 8.95	\\
$p_y$	&	19.81 / 31.97	&	13.32 / 18.5	&	17.94 / 22.59	&	7.23 / 10.22 \\
\bottomrule
\end{tabular}
\end{table}


Figure~\ref{fig:scratch_model_distribution} and Table~\ref{tab:model_statistics2} show
the parameters estimated for tabula rasa models that were trained from scratch. 
A comparison between these results and those of the previous section shows that the difference in results for the Jester and UCF101 datasets becomes more pronounced for tabula rasa models. This is to be expected since models trained from scratch are not endowed with good visual features and likely develop a larger repertoire of temporal parameters to compensate for this. It seems that, on the Jester dataset, the model needs to develop a larger variety of temporal parameters to do classification. This suggests that classification on the Jester dataset is more dependent on affine motion transformations than classification on the UCF101 dataset. This is not surprising since many of the classes in UCF101 are not dependent on unique motion patterns at all. For example, the SkyDiving, Skiing, and Skijet classes can be classified based on spatial features alone. In contrast, the classes in the Jester dataset are strongly dependent on what kind of motion the person performs with their hand. 

The comparison with pretrained models further reveals that the dependency of the models on the temporal parameters decreases strongly across models and datasets when initialization with pretrained weights is provided. This implies that when the model can extract good spatial features, the dependency on learning a broad range of temporal parameters decreases. This suggests that tabula rasa 3TConv models develop a greater dependency on temporal parameters. This interesting phenomenon warrants further investigation. 


\section{Conclusion}

Until now methods for explaining temporal features learned by 3DConvNets, in terms of simple and interpretable parameters, were not available. In this paper, we make an attempt to bridge the gap between the ever-increasing demand for deep learning models to be interpretable and the availability of methods that allow us to interpret said models. We introduced the 3TConv as an interpretable alternative to the regular 3DConv. By training different models on different datasets and analyzing the results, we achieve novel insight into what temporal aspects drive model classification. It is also demonstrated that 3TConvNets can make use of pretrained 2DConv weights in order to boost classification performance, in some cases surpassing the performance of traditional pretrained 3DConvs while containing up to 50\% less parameters. 


\newpage
\section*{Broader Impact}
This paper contributes to a more informed evaluation of the potential societal consequences of deep learning approaches to video analysis by proposing a method to achieve clarity about the nature of information processing in 3TConvNets. An important application of the type of research reported here is the usage of video classification for surveillance. This can have both positive and negative societal implications, depending on the purposes for which video classification is used, and on the effectiveness of the oversight mechanisms that aim to uphold conformity to the regulations, such as the GDPR in the EU. For a proper assessment of such implications, it is vital that human understanding of the computational processing involved is possible. The current research contributes to the explainability of 3D feature analysis by providing visual representations of what is being learned, in a format that is intuitively interpretable by humans. Therefore, it will be valuable to stakeholders that are interested in applying automated surveillance, as well as to stakeholders that are concerned about the potential infringement of human rights via such techniques.


\begin{ack}
We would like to thank Erdi \c{C}all{\i} and Elsbeth van Dam for the helpful discussions and general support.

\end{ack}


\bibliographystyle{vancouver}
\bibliography{ref}

\end{document}